%% file: manuscript.tex
\documentclass[journal]{IEEEtran}

\usepackage{amsmath,amsfonts}
\usepackage{algorithmic}
\usepackage[ruled,vlined]{algorithm2e}
\usepackage{array}
\usepackage{subfig}
\usepackage{textcomp}
\usepackage{stfloats}
\usepackage{url}
\usepackage{verbatim}
\usepackage{graphicx}
\usepackage{cite}
\usepackage{graphicx}
\def\BibTeX{{\rm B\kern-.05em{\sc i\kern-.025em b}\kern-.08em
    T\kern-.1667em\lower.7ex\hbox{E}\kern-.125emX}}
\usepackage{balance}
\begin{document}
\title{Compressed Particle-Based Federated Bayesian Learning and Unlearning}
\author{Jinu~Gong,~\IEEEmembership{Student Member,~IEEE,}
        Osvaldo~Simeone,~\IEEEmembership{Fellow,~IEEE,}
        and~Joonhyuk~Kang,~\IEEEmembership{Member,~IEEE}
\thanks{The work of J. Gong and J. Kang was supported by the MSIT (Ministry of Science and ICT), Korea, under the ITRC (Information Technology Research Center) support program (IITP-2020-0-01787) supervised by the IITP (Institute of Information \& Communications Technology Planning \& Evaluation). The work of O. Simeone was supported by the European Research Council (ERC) under the European Union's Horizon 2020 research and innovation programme (grant agreement No. 725731)}
\thanks{Jinu Gong and Joonhyuk Kang are with the School of Electrical Engineering, Korea Advanced Institute of Science and Technology, Daejeon, 34141 Korea (e-mail: kjw7419@kaist.ac.kr, jhkang@kaist.edu).}
\thanks{Osvaldo Simeone is with the Department of Engineering, King’s College London, London WC2R 2LS, U.K. (e-mail: osvaldo.simeone@kcl.ac.uk).}}

\markboth{}
{Shell \MakeLowercase{\textit{et al.}}: Bare Demo of IEEEtran.cls for IEEE Journals}

\maketitle

\begin{abstract}
Conventional frequentist FL schemes are known to yield overconfident decisions. Bayesian FL addresses this issue by allowing agents to process and exchange uncertainty information encoded in distributions over the model parameters. However, this comes at the cost of a larger per-iteration communication overhead. This letter investigates whether Bayesian FL can still provide advantages in terms of calibration when constraining communication bandwidth. We present compressed particle-based Bayesian FL protocols for FL and federated ``unlearning" that apply quantization and sparsification across multiple particles. The experimental results confirm that the benefits of Bayesian FL are robust to bandwidth constraints.
\end{abstract}

\begin{IEEEkeywords}
Federated learning, Bayesian learning, Stein variational gradient descent, Machine unlearning, Wireless communication
\end{IEEEkeywords}

\setlength{\abovedisplayskip}{3pt}
\setlength{\belowdisplayskip}{4pt}
\input{introduction}
\vspace{-0.1cm}

\input{problem}

\vspace{-0.1cm}

\input{algorithm}
\vspace{-0.1cm}

\input{experiments}
\vspace{-0.1cm}

\input{conclusion}

\bibliographystyle{IEEEtran}
\bibliography{manuscript}
\end{document}

%% file: introduction.tex
\section{Introduction}
\IEEEPARstart{D}{istributed} intelligence is envisaged to be one of the key use cases for 6G. An important primitive for the implementation of distributed intelligence is federated learning (FL), which supports distributed gradient-based training across a network of learning agents (see \cite{simeone2022machine} for an overview). Individual agents are often mobile devices with limited data and power \cite{kholod2020open,shyu2021systematic}. Despite such limitations, the decisions made by machine learning models trained via FL are expected to be used for sensitive applications such as personal healthcare. Furthermore, in such cases, agents may exercise their \textit{right to be forgotten}, requesting that information about their data be ``removed" from trained models available in the network for use by other devices \cite{ginart2019making}. This paper addresses the problem of developing communication-efficient FL protocols offering a reliable quantification of uncertainty while also supporting the right to erasure.

Most studies on FL are conducted within a frequentist framework, whereby agents perform local optimization in the space of model parameters, and iteratively exchange information about the updated model parameters through a server. Given the limited data available at each agent, there is uncertainty about the model parameters that are best suited to generalize outside the training set. By neglecting such uncertainty, frequentist learning schemes are known to yield poorly \textit{calibrated} decisions, which are typically overconfident \cite{guo2017calibration,khan2021bayesian}. Furthermore, in an FL framework, the ``collapse" of uncertainty in the model parameter space -- also known as epistemic uncertainty -- to a single model parameter vector prevents agents from properly communicating their respective states of knowledge about the problem. This, in turn, can yield slower convergence \cite{kassab2022federated}.

\begin{figure}[t]
    \centering
    \subfloat[Compressed particle-based federated Bayesian learning]{\includegraphics[width=0.48\textwidth]{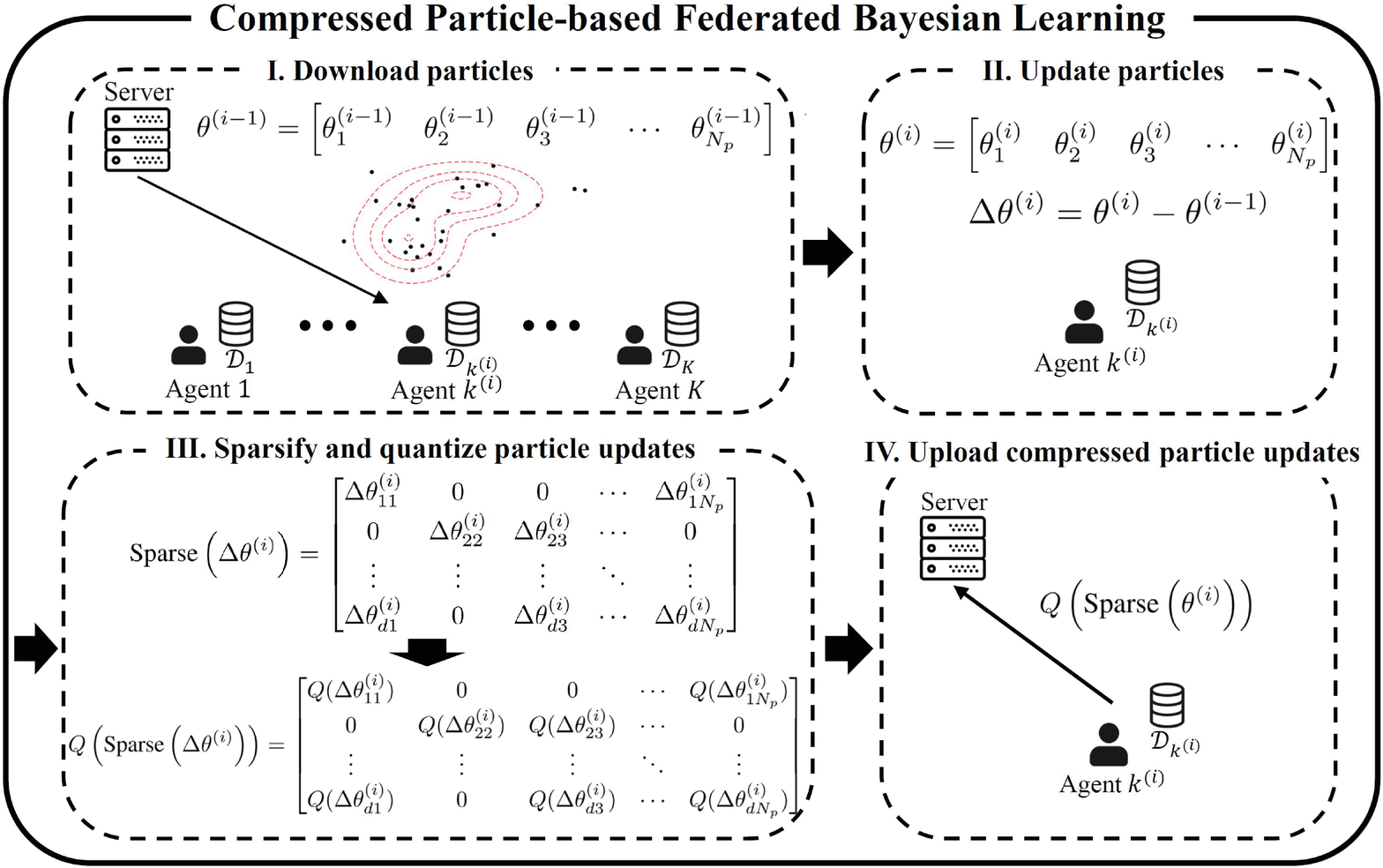}\label{fig:intro_learning}}
    \vspace{-0.2cm}\\
    \subfloat[Compressed particle-based federated Bayesian unlearning]{\includegraphics[width=0.48\textwidth]{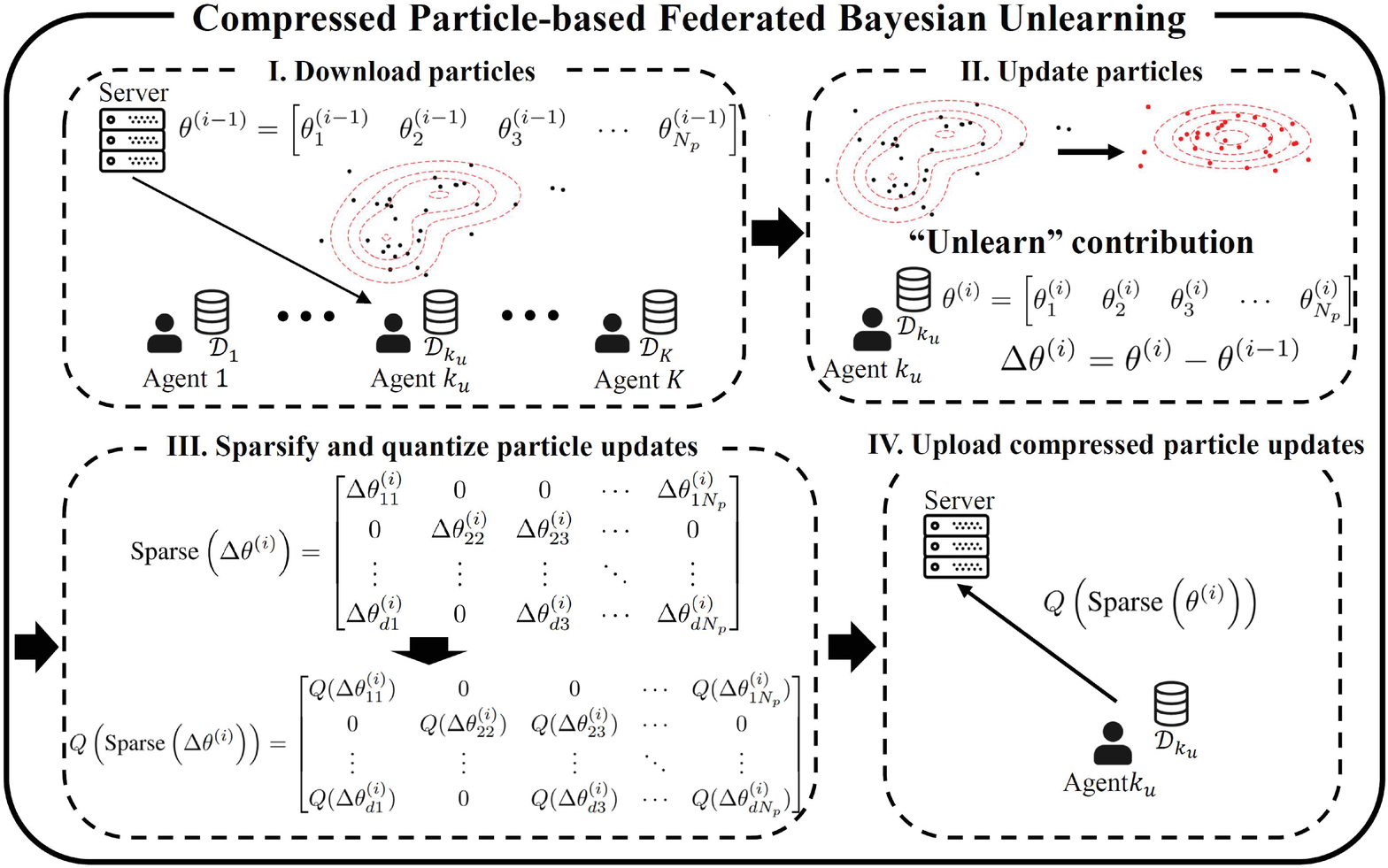}\label{fig:intro_unlearning}}\\
    \caption{Compressed Particle-based federated Bayesian learning and unlearning.}
    \vspace{-0.2cm}
    \label{fig:intro_figure}
\vspace{-0.4cm}
\end{figure}

A possible solution to this problem lies in adapting Bayesian learning methods, and generalizations thereof \cite{knoblauch2019generalized,simeone2022machine,jose2021free}, to FL. Bayesian learning optimizes probability distributions over the model parameter space, allowing for a representation of the state of epistemic uncertainty caused by limited data at the agents. Practical implementations of Bayesian learning represent the model parameter distribution either via a parametric family of distributions -- an approach known as variational inference (VI) -- or via a set of random particles -- following Monte Carlo (MC) sampling methods. MC-based methods can accurately estimate the target posterior distribution in the asymptotic regime of a large number of iterations, but they suffer from slow convergence. In contrast, VI-based methods have a significantly lower iteration complexity, but their performance is limited by the bias caused by the choice of a parametric family. 

\textit{Stein variational gradient descent} (SVGD) \cite{liu2016stein} is a non-parametric VI method that strikes a balance between expressivity of the approximation and iteration complexity. SVGD approximates the posterior distribution using a set of particles, like MC sampling, while also benefiting from the faster convergence of VI through deterministic optimization, rather than sampling. The \textit{Distributed SVGD} (DSVGD) protocol introduced in \cite{kassab2022federated} extends SVGD to FL (see Fig. \ref{fig:intro_learning}). The authors demonstrate the advantages of DSVGD in terms of the number of iterations and in terms of calibration with respect to standard frequentist FL. DSVGD was later adapted in \cite{gong2021forget} to introduce \textit{Forget-DSVGD}, a protocol that accommodates the right to erasure by leveraging the VI framework for machine unlearning presented in \cite{nguyen2020variational} (see Fig. \ref{fig:intro_unlearning}). 

Previous work \cite{kassab2022federated} has assumed the possibility to transfer an unlimited amount of information at each iteration round. Therefore, the advantages highlighted in \cite{kassab2022federated} of Bayesian FL were obtained at the cost of larger \textit{per-iteration} communication overhead. In fact, in DSVGD, agents need to exchange multiple particles at each iteration, rather than a single model parameter vector as in frequentist FL. This paper investigates the question of whether Bayesian FL can still provide advantages in terms of iteration complexity and calibration when constraining communication bandwidth between agents and server. To address this problem, we present compressed DSVGD and Forget-DSVGD, which apply quantization and sparsification across multiple particles. While quantization and sparsification have been widely applied to frequentist FL \cite{alistarh2017qsgd}, this is the first work to consider such techniques to reduce the communication overhead of particle-based Bayesian FL. 

The rest of this letter is organized as follows. Sec. \ref{sec:system} introduces the setup, and reviews the necessary background. Sec. \ref{sec:algorithm} presents the proposed compressed-DSVGD algorithm. Sec. \ref{sec:exp} describes numerical results, and Sec. \ref{sec:conclusion} concludes the paper.

%% file: problem.tex
\section{System Setup and Preliminaries}
\label{sec:system}
\vspace{-0.1cm}
\subsection{Setup}
As illustrated in Fig. \ref{fig:intro_figure}, we consider a federated learning setup with a set $\mathcal{K}=\{1,\ldots,K\}$ of $K$ agents within a parameter-server architecture (see, e.g., \cite{simeone2022machine}). The local data set $\mathcal{D}_{k}=\{z_{k,n} \}_{n=1}^{N_k}$ of agent $k\in\mathcal{K}$ contains $N_k$ data points. The collection of all local datasets is referred to as the global data set $\mathcal{D}$. We express the local training loss at agent $k$ with respect to the $d\times 1$ model parameter $\theta$ as the empirical average
\begin{align}
L_k(\theta)=\frac{1}{N_k}\sum_{n=1}^{N_k} \ell (z_{k,n}|\theta)
\end{align}
for some loss function $\ell (z|\theta)$. For a likelihood function $p(z|\theta)$, the loss function is typically chosen as the log-loss $\ell (z|\theta)=-\log p(z|\theta)$.

In Bayesian federated learning, the goal is to obtain a \textit{variational distribution} $q(\theta)$ on the model parameter space that minimizes the \textit{global free energy} (see, e.g., \cite{kassab2022federated,bui2018partitioned,simeone2022machine})
\begin{align}
\min_{q(\theta)} \bigg\{ \!F(q(\theta))\!=\!\sum_{k=1 }^K& N_k \mathbb{E}_{\theta\sim q(\theta)}[L_k(\theta)]+\!\alpha \!\cdot \!\mathbb{D}\big( q(\theta)\big\|p_0 (\theta)\big) \bigg\},\label{eq:global_fe}
\end{align}
where $\alpha > 0$ is a ``temperature" parameter; $\mathbb{D}\left(\cdot\|\cdot\right)$ is the Kullback–Leibler (KL) divergence; and $p_0(\theta)$ denotes a prior distribution. The optimization problem (\ref{eq:global_fe}) seeks for a distribution $q(\theta)$ that minimizes the average sum-training loss, i.e., the first term in (\ref{eq:global_fe}), while being close to the prior distribution $p_0 (\theta)$, as enforced by the second term.

The unconstrained optimal solution of problem (\ref{eq:global_fe}) is given by the \textit{global generalized posterior distribution}
\begin{align}
q^*(\theta|\mathcal{D})&=\frac{1}{Z}\cdot \tilde{q}^*(\theta|\mathcal{D})\label{eq:opt_sol_ori}\\ \textrm{where}\quad\tilde{q}^*(\theta|\mathcal{D})&=p_0 (\theta) \exp \left(-\frac{1}{\alpha}\sum_{k=1}^K N_k L_k(\theta)\right),\label{eq:opt_sol}
\end{align}
which equals the conventional posterior distribution $p\big(\theta|\mathcal{D}\big)$ when one sets $\alpha=1$ and the loss function as the \textit{log-loss} $\ell(z|\theta)=-\log p(z|\theta)$.

However, in practice, problem (\ref{eq:global_fe}) can only be solved in an approximate manner by using parametric or non-parametric methods. In this letter, we focus on a state-of-the-art non-parametric particle-based method, SVGD \cite{liu2016stein}, which represents the distribution $q(\theta)$ in (\ref{eq:global_fe}) in terms of $N_p$ particles $\{ \theta_1, \ldots, \theta_{N_p} \}$ (see Fig. \ref{fig:intro_figure}). Given particles $\{ \theta_1, \ldots, \theta_{N_p} \}$, an explicit estimate of distribution $q (\theta)$ can be obtained, e.g., via kernel density estimator (KDE) with some kernel function $K(\theta, \theta')$, i.e., $q(\theta)=\frac{1}{N_p}\sum_{n=1}^{N_p} K(\theta, \theta_n)$ (see, e.g., \cite{simeone2022machine}).

\vspace{-0.1cm}
\subsection{Distributed SVGD}
DSVGD addresses problem (\ref{eq:global_fe}) in a federated setting by describing distribution $q(\theta)$ via a set of $N_p$ particles $\{\theta_n \}_{n=1}^{N_p}$ that are updated by scheduling a subset of agents each iteration (see, e.g., \cite{kassab2022federated}). In this letter, we focus on the case of a single agent scheduled at each iteration, since the extension to more than one agent is direct by following the approach in \cite{kassab2022federated}.

At the beginning of the $i$-th iteration, the server stores the current global particles $\{\theta_n^{(i-1)} \}_{n=1}^{N_p}$, which represent the current iterate $q^{(i-1)}(\theta)$ of the global variational distribution. The variational distribution $q^{(i-1)}(\theta)$ is modelled via the factorization $q^{(i-1)}(\theta)=p_0(\theta)\prod_{k=1}^K t_k^{(i-1)}(\theta)$ \cite{ bui2018partitioned,kassab2022federated,simeone2022machine}, where the term $t_k^{(i-1)} (\theta)$ is known as approximate likelihood of agent $k$. At each iteration $i$, the scheduled agent $k^{(i)}$ updates the variational distribution $q^{(i-1)}(\theta)$ by modifying its approximate likelihood to a new iterate $t_{k^{(i)}}^{(i)}(\theta)$ via the optimization of a set of local particles. Specifically, given kernel functions $K(\cdot,\cdot)$ and $\kappa (\cdot, \cdot)$, DSVGD operate as follows \cite{kassab2022federated}.
\noindent\textbf{Initialization.} Draw the set of $N_p$ global particles $\{\theta_n^{(0)} \}_{n=1}^{N_p}$ from the prior $p_0(\theta)$; and initialize at random the set of local particles $\{ \theta_{k,n}^{(0)}  \}_{n=1}^{N_p}$for all agents $k\in\mathcal{K}$.

\noindent\textbf{Step 1.} At each iteration $i$, server schedules an agent $k^{(i)}\in\mathcal{K}$. Agent $k^{(i)}$ downloads the current global particles $\{\theta_n^{(i-1)} \}_{n=1}^{N_p}$ from the server.

\noindent\textbf{Step 2.} Agent $k^{(i)}$ initializes its particles to equal the global particles, i.e., $\{\theta_{n}^{[0]}=\theta_{n}^{(i-1)}\}_{n=1}^{N_p}$. Furthermore, it sets its local likelihood to $t_{k^{(i)}}^{(i-1)}(\theta)= {1}/{N_p} \sum_{n=1}^{N_p} K(\theta, \theta_{{k^{(i)}},n}^{(i-1)})$ and the global posterior to $q^{(i-1)}(\theta)= {1}/{N_p} \sum_{n=1}^{N_p} K(\theta, \theta_n^{(i-1)})$. Then, it updates the particles via SVGD \cite{kassab2022federated} as
\begin{align}
\theta_n^{[l]}&\leftarrow\theta_{n}^{[l-1]}+\epsilon \phi\left({\theta_n^{[l-1]}}\right),\label{eq:unlearn_svgd_particle}
\end{align}
for all particles $n=1,\ldots, {N_p}$ with learning rate $\epsilon$, and function 
\begin{align}
\phi(\theta)=\frac{1}{{N_p}}\sum_{j=1}^{N_p} \bigg[\kappa\left(\theta_j^{[l-1]},\theta\right)&\nabla_{\theta_j}\log \tilde{p}_{k^{(i)}}^{(i)}\left(\theta_j^{[l-1]}\right)\nonumber\\
&+\nabla_{\theta_j}\kappa\left(\theta_j^{[l-1]},\theta\right)\bigg],\label{eq:phi_hat}
\end{align}
across local iterations $l=1,\ldots,L$, where we have defined the ``tilted" distribution as
\begin{align}
\tilde{p}_{k^{(i)}}^{(i)}(\theta)\propto \frac{q^{(i-1)}(\theta)}{t_{k^{(i)}}^{(i-1)}(\theta)} \exp \left(-\frac{1}{\alpha} L_{k^{(i)}}(\theta)\right). \label{eq:tilted_dist}
\end{align}
\noindent\textbf{Step 3.} After $L$ local iteration, agent $k^{(i)}$ sets $\{\theta_n^{(i)}=\theta_n^{[L]}\}_{n=1}^{N_p}$. The updated global particles $\{\theta_n^{(i)} \}_{n=1}^{N_p}$ are sent to the server, which sets $\{\theta_n =\theta_n^{(i)} \}_{n=1}^{N_p}$. Finally, agent $k^{(i)}$ updates its local particles $\{\theta_{{k^{(i)}}, n}^{(i)} \}_{n=1}^{N_p}$ using the updated global particles $\{\theta_n^{(i)} \}_{n=1}^{N_p}$, while the other agents $k'\neq {k^{(i)}}$ set $\{\theta_{{k}',n}^{(i)} =\theta_{k',n}^{(i-1)} \}_{n=1}^{N_p}$. We refer to \cite[Sec. 5.2]{kassab2022federated} for benefits on the update of the local particles.

\subsection{Forget-SVGD}

We finally describe the variational unlearning formulation in \cite{nguyen2020variational}, which is referred to as Forget-SVGD. Before unlearning, Forget-SVGD assumes that an approximate solution $q(\theta|\mathcal{D})$ of the federated learning problem (\ref{eq:global_fe}) has been obtained, e.g., via DSVGD. Forget-SVGD aims at removing the contribution for data of a subset $\mathcal{U}\subset \mathcal{K}$ of agents, which wish to unlearn, from the learned model $q(\theta|\mathcal{D})$.

A baseline approach would retrain \textit{from scratch} the global model excluding the agents in subset $\mathcal{U}$. A potentially more efficient solution, Forget-SVGD, operates as follows \cite{gong2021forget}.

\noindent\textbf{Initialization.} The initial set of $N_p$ particles $\{\theta_n^{(0)}\}_{n=1}^{N_p}$ represents the variational distribution obtained as a result of Bayesian federated learning; initialize at random local particles $\{\theta_{k,n}^{(0)} \}_{n=1}^{N_p}$ for all agents $k\in \mathcal{U}$

\noindent\textbf{Step 1.} At iteration $i$, the server schedules an agent ${k^{(i)}}\in\mathcal{U}$, within the set of agents who have requested their data to be ``forgotten". Agent ${k^{(i)}}$ downloads the current global particles $\{\theta_n^{(i-1)} \}_{n=1}^{N_p}$ from the server.

\noindent\textbf{Step 2.} Agent ${k^{(i)}}$ initializes the particles $\{\theta_{n}^{[0]}=\theta_{n}^{(i-1)}\}_{n=1}^{N_p}$, and it updates the particles using the SVGD update (\ref{eq:unlearn_svgd_particle})-(\ref{eq:phi_hat}) by replacing the tilted distribution in (\ref{eq:tilted_dist}) with 
\begin{align}
\tilde{p}_{k^{(i)}}^{(i)}(\theta)=\frac{q^{(i-1)}(\theta)}{t_{k^{(i)}}^{(i-1)}(\theta)} \exp \left(\frac{1}{\alpha} L_{k^{(i)}}(\theta)\right)\label{eq:tilted_dist_ul},
\end{align}
where $q^{(i-1)}(\theta)$ and $t_{k^{(i)}}^{(i-1)}(\theta)$ are computed by using the respective KDEs with global and local particles, respectively.

\noindent\textbf{Step 3.} This step applies the same operations as Step 3 of DSVGD.

%% file: algorithm.tex
\section{Compressed-Distributed Stein Variational Gradient Descent}
\label{sec:algorithm}
In this section, we study an implementation of DSVGD and Forget-SVGD over rate-constrained channels between agents and server. This constraint affects the uploading of the updated global particles $\{\theta_n^{(i)} \}_{n=1}^{N_p}$ at each $i$-th iteration of Step 3 of DSVGD and Forget-SVGD. Downlink communication from server to agents is assumed to be noiseless in order to focus on the more challenging uplink channel. Accordingly, the scheduled agent $k^{(i)}$ can communicate no more than $R_u$ bits per iteration.

To facilitate compression, at each $i$-th iteration, agent $k^{(i)}$ uploads the $N_p\times d$ matrix $\Delta\Theta^{(i)}=\left[\theta_1^{(i)}-\theta_1^{(i-1)}, \theta_2^{(i)}-\theta_2^{(i-1)}, \ldots, \theta_{N_p}^{(i)}-\theta_{N_p}^{(i-1)}\right]$ of updates for all particles to the server. This is a common step in compressed frequentist FL algorithms, which communicate a single parameter vector per iteration, i.e., they have $N_p=1$ \cite{alistarh2017qsgd}. In this section, we develop compression strategies based on sparsification and quantization of the updates $\Delta\Theta^{(i)}$ in order to meet the capacity constraint of $R_u$ bits per iteration. The key novel element as compared to prior work is the need to compress multiple particles simultaneously.

\subsection{Sparsification}
Top-$k$ sparsification is a widely used method for frequentist FL that selects the entries of the model parameter vector with the $k$ largest absolute values. All other entries of the update vector are set to zero. In this work, we introduce and study the following variants of top-$k$ sparsification for particle-based Bayesian learning based on DSVGD. For all schemes, to identify the sparsified position, we assume Golomb position encoding, which requires $\log_2 \binom{d}{k}$ for an input vector of $d$ entries \cite{sattler2019sparse}. We denote as $N_b$ the number of bits used to represent each entry retained by the sparsification process.

\begin{figure}[t]
\centering
\includegraphics[width=8.3cm]{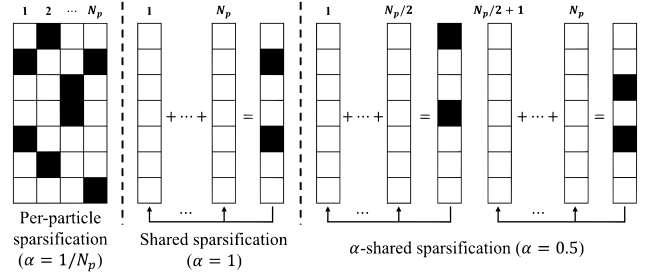}
\caption{Sparsification methods.}
\label{fig:spars_method}
\vspace{-0.4cm}
\end{figure}

\subsubsection{Per-particle sparsification}
A baseline approach is to apply top-$k$ sparsification separately to each particle update (see Fig. \ref{fig:spars_method}-(left)). This scheme requires
\begin{align}
R_u=N_p\times \left( \log_2{ \binom{d}{{ r\times d}} }+N_b \times {r\times d} \right)
\end{align}
bits per iteration, where we have defined the per-particle sparsification ratio $r=k/d$, and the first term in the parenthesis is the number of bits required to specify the top-$k$ positions in each particle. The communication overhead necessary to identify the top-$k$ entries hence scales linearly with the number $N_p$ of particles.
\subsubsection{Shared sparsification}
When the bit rate $R_u$ is small, a potentially more efficient approach is based on the assumption that the sparsity pattern is common to all particles. To implement this idea, which we refer to as shared sparsification, we sum the absolute values of each entry of the $N_p$ particles, and the top-$k$ entries are selected based on the resulting sum vector. The resulting sparsity pattern is  applied to all particles (see Fig. \ref{fig:spars_method}-(center)). This scheme requires 
\begin{align}
R_u = \log_2{ \binom{d}{{ r\times d }} }+ N_p \times N_b \times {r\times d}   
\end{align}
bits per iteration, reducing by $N_p$ times the overhead for position encoding.
\subsubsection{$\alpha$-shared sparsification}
Generalizing the previous two schemes, $\alpha$-shared sparsification divides the particles into $1/\alpha$ groups, and only shares the sparsity pattern among particles in the same group. For each group, the scheme applies the same procedure of shared sparsification method (see Fig. \ref{fig:spars_method}-(right)). Note that setting $\alpha=1/N_p$ yields per-particle sparsification; and setting $\alpha=1$ yields shared sparsity. More generally, the scheme is defined for every value $\alpha \in [1/N_p,1]$ such that $1/\alpha$ is an integer that divides $N_p$. This scheme requires 
\begin{align}
R_u = \frac{1}{\alpha}\times\log_2{ \binom{d}{{ r\times d }} }+ N_p \times N_b \times {r\times d}
\end{align}
bits per iteration, reducing by $N_p$ times the overhead for position encoding.


\subsection{Quantization}\label{sec:quantiazation}
Every entry selected by the sparsification step is finally quantized using stochastic quantization \cite{alistarh2017qsgd}. For each entry $x\in \mathbb{R}$, the scheme requires 1 bit for the sign $\textrm{sign}(x)$, and $N_b -1$ bits for the magnitude $|x|$. Within a predefined dynamic range $[0, a_{\max}]$, a step size $\delta={a_{\max}}/(2^{N_b}-1)$ is set, and the stochastic quantizer $Q_{N_b} (x)$ is defined as
\begin{align}
Q_{N_b} (x)= \textrm{sign}(x)\cdot \zeta(\textrm{clip}(|x|),N_b),
\end{align}
where
\begin{align}
\zeta(a,N_b)=\begin{cases}
t \delta &\textrm{with probability } 1-\frac{a-t\delta}{\delta}\\
(t+1) \delta &\textrm{otherwise},
\end{cases}
\end{align}
and $\textrm{clip}(a)=\min (a, a_{\max})$.

%% file: experiments.tex
\section{Experiments}
\label{sec:exp}
\subsection{Federated Learning}
We are interested in comparing the performance of frequentist FL and Bayesian FL in the presence of an uplink per-iteration rate constraint $R_u$. For frequentist FL, we adopt FedAvg with standard top-$k$ sparsification and stochastic quantization as in, e.g., \cite{alistarh2017qsgd, qin2021federated}. We have $K=10$ agents, each with $N_k=6000$ examples from the Fashion-MNIST data set. The model consists of one fully-connected hidden layer with 100 hidden neurons and a softmax output layer. For compressed-DSVGD, as in \cite{liu2016stein}, we consider the radial basis function (RBF) kernel $\kappa (x,x')=\textrm{exp}(-\| x-x' \|_{2}^2 /h)$ and the bandwidth $h=\textrm{med}^2/\log N$, where $\textrm{med}$ is the median of the pairwise distances between the particles in the current iteration. We also assume the Gaussian kernel $K(x,x')\propto \exp (-\|x-x'\|^2/\lambda)$ for the KDE with a bandwidth $\lambda=0.55$. The fixed temperature parameter is set to $\alpha=1$, and AdaGrad \cite{liu2016stein} is used to determine the learning rate schedule in (\ref{eq:unlearn_svgd_particle}).

\begin{figure}[t]
\centering
\includegraphics[width=8.7cm]{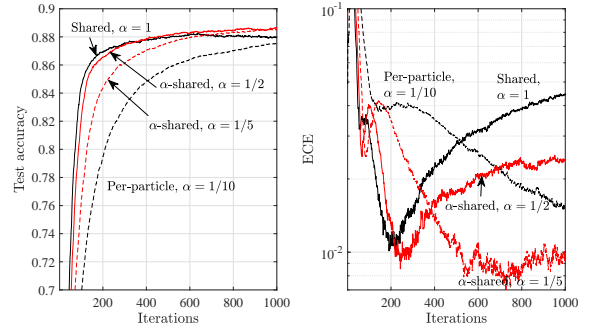}
\caption{Test accuracy and ECE for three sparsification methods with $R_u=d$ bits per iteration (Fashion-MNIST data set, $K=10$ devices).}
\label{fig:figure2}
\vspace{-0.4cm}
\end{figure}

We evaluate the performance by using two metrics, namely test accuracy and \textit{expected calibration error} (ECE) \cite{guo2017calibration}. The ECE measures the capacity of a model to quantify uncertainty. It does so by evaluating the difference between the confidence level output by the model and the actual test accuracy. The confidence level is given by the output of the last, softmax, layer corresponding to the prediction of the model. The ECE is defined by partitioning the test set into $M$ bins $\{B_m \}_{m=1}^M$ depending on the confidence level of the model's decision, and by evaluating the accuracy $\textrm{acc}\left( B_m \right)$ for the examples within each bin. The ECE is given by the average of the difference between accuracy $\textrm{acc}\left( B_m \right)$ and confidence $\textrm{conf}\left( B_m \right)$ across all bins as \cite{guo2017calibration}
\begin{align}
\textrm{ECE}=\sum_{m=1}^M \frac{|B_m|}{n} \left| \textrm{acc}\left( B_m \right) - \textrm{conf}\left( B_m \right)  \right|,
\end{align}
where $|B_m|$ is the number of test examples in the $m$-th bin.

We start by comparing the performance of compressed-DSVGD under the proposed sparsification methods by setting the number of particle to $N_p=10$, the number of quantization bits to $N_b=5$, and the per-iteration rate to $R_u=d$. Fig. \ref{fig:figure2} plots test accuracy and ECE as a function of the training iterations, where average results are reported over $10^2$ runs of the algorithms. The figure suggests that, while shared sparsification is most effective when we can only run a small number of iterations, shared sparsification with $\alpha<1$ is required to obtain smaller values of test error and ECE. Note that the minimum value of $\alpha$, $\alpha=1/10$, which corresponds to per-particle sparsification is generally suboptimal.
\begin{figure}[t]
\centering
\includegraphics[width=7.5cm]{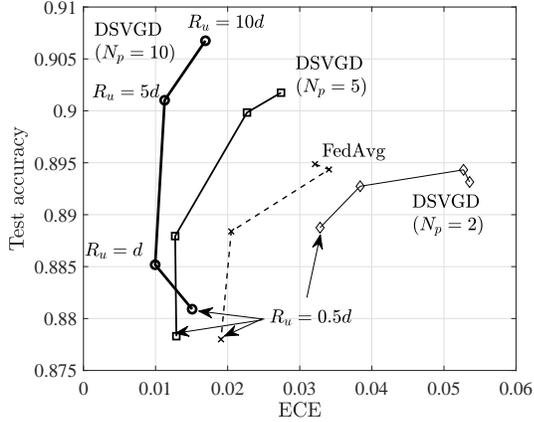}
\caption{Test accuracy and ECE plot for FedAvg and DSVGD with $N_p=2, 5, 10$ particles under per-iteration bit constraints $R_u=0.5d, d, 5d, 10d$. The markers indicate points with the same value of $R_u$.}
\label{fig:learning}
\vspace{-0.4cm}
\end{figure}

In Fig. \ref{fig:learning}, we present test accuracy with respect to ECE after $10^3$ training iterations for FedAvg and DSVGD with $N_p=2, 5, 10$ particles. We apply $\alpha$-shared sparsity and vary the per-iteration bits constraints $R_u=0.5d, d, 5d, 10d$. As shown in Fig. \ref{fig:learning}, DSVGD outperforms FedAvg in test accuracy and ECE, even under the per-iteration bit constraints, unless the number of particles, $N_p$, is too low, here $N_p=2$. Furthermore, as a per-iteration bits constraint $R_u$ decreases, it is more beneficial to reduce the number of particles $N_p$ than to decrease the sparsification ratio $r$.

\subsection{Federated Unlearning}
\begin{figure}[t]
\centering
\includegraphics[width=8.7cm]{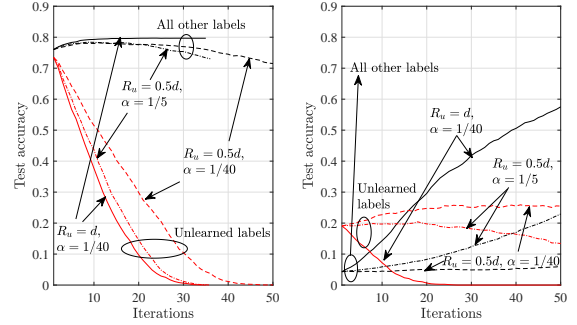}
\caption{Average test accuracy of the unlearned labels and of the remaining labels for unlearning compressed-Forget-SVGD and the training from scratch using the data set $\mathcal{D}\setminus \mathcal{D}_u$ via compressed-DSVGD under per bit constraints $R_u= d,\,0.5d$.}
\label{fig:unlearning}
\vspace{-0.4cm}
\end{figure}

For federated unlearning, we adopt a ``non-iid" setting with $K=10$ agents by assigning each agent $100$ examples from only two of the ten classes of Fashion-MNIST images. The two agents with labels 2 and 9 request that their contribution be ``unlearned". We follow references \cite{kristiadi2020being, jospin2022hands} by pre-training using conventional FedAvg, and then training the last layer using DSVGD with $N_p=40$ particles. Then, we ``unlearn" the model based on the proposed compressed Forget-SVGD scheme. Fig. \ref{fig:unlearning}-(left) shows the average test accuracy for the unlearned labels (2 and 9) and that of remaining labels during compressed-Forget-SVGD iterations for per-iteration bit constraints $R_u= d, 0.5 d$. The right panel shows, for reference, the performance of a train-from-scratch scheme using only the remaining labels, which is seen to be significantly slower. For a smaller bandwidth $R_u$, here $R_u= 0.5d$, using a larger $\alpha$ tends to degrade, as desired, the accuracy for the unlearned labels, while also affecting the performance of the other labels. This points to a trade-off between forgetting and retraining useful information that can be controlled via the parameter $\alpha$.

%% file: conclusion.tex
\section{Conclusion}
\label{sec:conclusion}
This letter has investigated the performance of particle-based Bayesian federated learning and unlearning under bandwidth constraints. A new class of sparsification methods was proposed that operates across multiple particles. Through simulations, we have confirmed that Bayesian FL can outperform standard frequentist FL in terms of test accuracy and calibration even under per-iteration bit constraints. Furthermore, we have identified a trade-off between forgetting requested data and retraining useful information that can be controlled by the choice of the sparsification scheme.